\documentclass{article}

\usepackage{microtype}
\usepackage{graphicx}
\usepackage{subcaption}
\usepackage{booktabs} 

\usepackage{hyperref}

\usepackage[preprint]{icml2026}

\usepackage{amsmath}
\usepackage{amssymb}
\usepackage{mathtools}
\usepackage{amsthm}
\usepackage{subfiles}
\usepackage{enumitem}
\usepackage{multirow}
\usepackage{xcolor}  
\usepackage{pifont}
\newcommand{\cmark}{\ding{51}}
\newcommand{\xmark}{\ding{55}}

\definecolor{checkmark_green}{RGB}{0, 100, 0}        
\definecolor{xmark_red}{RGB}{178, 34, 34}            

\setlist{itemsep=2pt, topsep=3pt, parsep=0pt, partopsep=0pt}
\newcommand{\proposed}{T\textsuperscript{3}VF}
\usepackage[capitalize,noabbrev]{cleveref}

\theoremstyle{plain}

\theoremstyle{definition}

\theoremstyle{remark}

\usepackage[textsize=tiny]{todonotes}
\icmltitlerunning{Test-Time Training for Visual Foresight Vision-Language-Action Models}
\begin{document}
\twocolumn[
      \icmltitle{Test-Time Training for Visual Foresight Vision-Language-Action Models}




  \begin{icmlauthorlist}
    \icmlauthor{Sangwu Park}{aaa}
    \icmlauthor{Wonjoong Kim}{aaa}
    \icmlauthor{Yeonjun In}{aaa}
    \icmlauthor{Sein Kim}{aaa}
    \icmlauthor{Hongseok Kang}{aaa}
    \icmlauthor{Chanyoung Park}{aaa}
  \end{icmlauthorlist}

  \icmlaffiliation{aaa}{KAIST, Seoul, South Korea}

  \icmlcorrespondingauthor{Chanyoung Park}{cy.park@kaist.ac.kr}

  \icmlkeywords{Vision-Language-Action Models, Test-Time Training, Out-of-Distribution}

  \vskip 0.3in
]



\printAffiliationsAndNotice{}  
\begin{abstract} \label{sec:abstract}
Visual Foresight VLA (VF-VLA) has become a prominent architectural choice in the recent VLA due to its impressive performance. Nevertheless, the inherent design of VF-VLA makes it particularly vulnerable to out-of-distribution (OOD) shifts. Because the quality of action directly depends on the accuracy of the predicted future visual information, OOD conditions affect both stages at once. To address this vulnerability, we propose Test-Time Training Visual Foresight VLA (\proposed), a test-time training approach motivated by the observation that the predicted future image and its subsequent observation form a natural supervision pair. To further address the practical challenges that arise from indiscriminate test-time updates, we introduce an adaptive update filtering mechanism. Empirically, \proposed~mitigates the OOD vulnerability of VF-VLA at a modest additional inference cost, without requiring any architectural modification or auxiliary modules\footnote{Our source code is available at \url{https://github.com/sangwu99/T3VF.git}.}.

\end{abstract}
\section{Introduction} \label{sec:introduction}
\vspace{-0.5ex}
Vision-Language-Action (VLA) models have become a central paradigm for generalist robotic manipulation \citep{kim2025fine, pertsch2025fast, bjorck2025gr00t}. Among them, a recent line of research adopts a two-stage formulation in which the model first predicts the future visual state that the robot is expected to reach and then generates actions conditioned on this prediction. Models that adopt this formulation are called Visual Foresight VLA (VF-VLA), and their impressive performance has made this design one of the prominent architectural choices in the recent VLA literature \citep{zhao2025cot, zhang2025up, wang2025unified, cen2025worldvla, zhangdreamvla, yang2025mantis}. 

\begin{figure}[t!]
\centering
    \includegraphics[width=\linewidth]{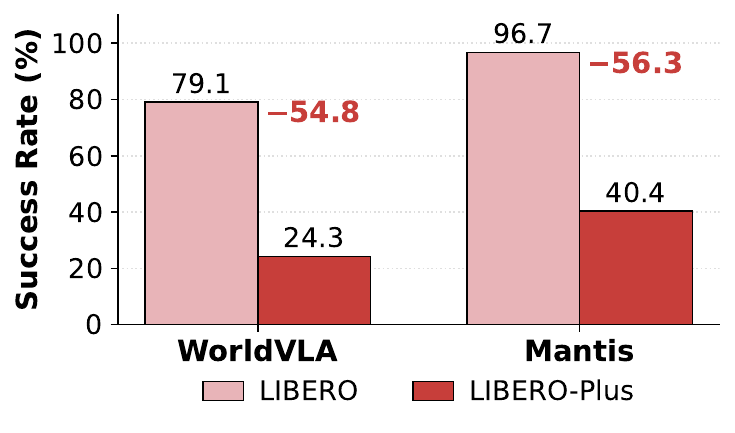}
    \caption{Performance comparison on LIBERO (In-Distribution) and LIBERO-Plus (Out-of-Distribution)}
    \label{fig:motivation}
\end{figure}

Despite the effectiveness of VF-VLA, its inherent design makes it particularly vulnerable to out-of-distribution (OOD) shifts. Because the quality of action generation directly depends on the accuracy of the predicted future visual information \citep{zhao2025cot}, OOD conditions affect both the visual prediction stage and the action generation stage at once. As summarized in Fig.~\ref{fig:motivation}, recent VF-VLA such as WorldVLA \citep{cen2025worldvla} and Mantis \citep{yang2025mantis} show substantial performance drops on LIBERO-Plus \citep{fei2025libero} compared to their in-distribution success rates on LIBERO \citep{liu2023libero}. This support the dual-stage exposure of both pathways under OOD shifts.


To address this vulnerability, we propose \textbf{T}est-\textbf{T}ime \textbf{T}raining \textbf{V}isual \textbf{F}oresight VLA (\proposed~), a test-time training approach motivated by a simple intuition that the predicted future image and its later observation form a natural self-supervision pair. In VF-VLA, at a given step the model predicts a future image from the current observation, and after executing the corresponding action, the actual image at that future step is observed. This observed image can be regarded as an oracle for the earlier prediction, allowing the test-time training to improve visual prediction. Accordingly, action generation also improves through the dependence.


However, this approach faces two practical challenges. First, a discrepancy between a prediction and its oracle does not always indicate a useful supervision signal. Such discrepancy can result from an inaccurate visual prediction or from an action-side error despite a correct prediction, which serves as noise during training. Second, the difficulty varies both across episodes and within an episode. Even with a proxy metric that can separate the two error sources discussed above, applying a fixed threshold cannot reflect this variation. Under such threshold, most steps in easier episodes or segments pass and allow excessive noise to enter training. In contrast, most steps in harder ones are skipped and fail to provide training signal.


To address the first challenge, we use the variance of the action at each step as the proxy metric. A low variance indicates that the model is internally consistent about the action it intends to execute. Such consistency makes a prediction error more plausibly attributable to the visual pathway and therefore a useful signal for updating the model. In contrast, high variance indicates that the source of the error cannot be reliably attributed. Since such an error may not provide a useful supervision signal, the corresponding step is skipped. To address the second challenge, we replace the fixed threshold with an adaptive variance buffer that maintains a running window of recent variances and permits an update only when its variance falls within the lower range of this buffer. Because this criterion is based on relative ranking rather than an absolute cutoff, it adapts naturally to the scale of variance in each episode as well as to its fluctuation between steps.

The main contributions of this work are as follows:
\begin{itemize}[leftmargin=.1in]
\vspace{-1ex}
\item \textbf{First identification of OOD vulnerability in VF-VLA.} We identify the amplified OOD vulnerability arising from the dual-stage exposure inherent to VF-VLA.
\item \textbf{Test-time training with predicted-attained pairs.} We propose a test-time training approach leverages the correspondence between predicted images and their later observations, complemented by an adaptive update filtering.
\item \textbf{Empirical effectiveness.} \proposed~improves the average success rate on LIBERO-Plus by 5\% over the base VF-VLA, demonstrating that it mitigates the OOD vulnerability.
\end{itemize}

\vspace{-1.5ex}
\section{Related Works} \label{sec:related_work}
\vspace{-0.5ex}
Visual Foresight VLA refer to VLA models that predict a future image to be reached by the robot from the current observation and language instruction, and condition action generation on this prediction. The first group formulates future visual pixels and actions as autoregressive token sequences on a shared vocabulary \citep{zhao2025cot, zhang2025up, wang2025unified, cen2025worldvla}. The second group predicts future visual information as compressed latent features or implicitly aligns latent representations with future visual states, avoiding the redundancy and computational overhead of autoregressive pixel-token generation. \citep{yang2025mantis, zhangdreamvla}. Both groups share a common structure: action generation is conditioned on predicted visual information. Consequently, both the visual prediction stage and the action generation stage are exposed to OOD shifts \citep{pumacay2024colosseum, fei2025libero}, leading to amplified vulnerability under such shifts. To the best of our knowledge, this vulnerability of VF-VLA remains unaddressed.


Another line of work adapts VLA models through test-time reinforcement learning. \citep{bai2025evolve, liu2026fly}. They differ from ours in that they require a separate reward model, incur substantial overhead from online RL, and target VLA models in general rather than VF-VLA. For complete related work, please refer to Appendix~\ref{app:related_work}.
\vspace{-1.5ex}
\section{Method} \label{sec:method}

\vspace{-0.5ex}
\subsection{Preliminaries}
\vspace{-0.5ex}
A VF-VLA consists of a VLM backbone $P$, an image head $I_h$, and an action head $A_h$. Given the language instruction $l$, the current observation $o_t$, and a placeholder $q$ for any learnable query tokens,\footnote{\label{fn:impl}The specific forms of placeholder $q$, $\mathcal{L}_{\mathrm{img}}$ and $\mathcal{L}_{\mathrm{act}}$ depend on the design choices of each VF-VLA implementation.} backbone extracts following representations
\begin{equation}
(h_t^{\mathrm{inst}},\, h_t^{\mathrm{img}},\, h_t^{\mathrm{act}}) \;=\; P(l,\, o_t,\, q).
\end{equation}
This step computes $h_t^{\mathrm{act}}$ conditionally on $h_t^{\mathrm{img}}$ following the autoregressive nature of the VLM backbone, which discussed in Sec.~\ref{sec:introduction}.
The image head predicts future observation $\hat{o}_{t+n}$ at a fixed gap of $n$ steps ahead, and the action head generates the action $\hat{a}_t$,
\begin{equation}
\hat{o}_{t+n} \;=\; I_h\!\left([h_t^{\mathrm{inst}}, h_t^{\mathrm{img}}],\, o_t\right), \quad \hat{a}_t \;\sim\; A_h(h_t^{\mathrm{act}}).
\end{equation}
During training, the model is optimized through two objectives\textsuperscript{\ref{fn:impl}},
\begin{equation}{\label{eqa:3}}
\mathcal{L}_{\mathrm{train}} \;=\; \mathcal{L}_{\mathrm{img}}(\hat{o}_{t+n},\, o_{t+n}) \;+\; \lambda \,\mathcal{L}_{\mathrm{act}}(\hat{a}_t,\, a_t),
\end{equation}
where $o_{t+n}$ is the future ground-truth observation, $a_t$ is the demonstration action, and $\lambda$ balances the two terms. 

\vspace{-0.5ex}
\subsection{Test-Time Training with Predicted-Attained Image Correspondence}\label{subsec:ttt}
\vspace{-0.5ex}
\textbf{Intuition.} At step $t$, the model predicts $\hat{o}_{t+n}$ and executes the corresponding action $\hat{a}_t$. After $n$ steps, the actual observation $o_{t+n}$ is attained from the environment. The predicted-attained pair $(\hat{o}_{t+n},\, o_{t+n})$ thus provides supervision for the image head in the same form as Eq.~\ref{eqa:3}, enabling test-time training without any additional data collection.


\begin{figure}[!t]
\centering
    \includegraphics[width=\linewidth]{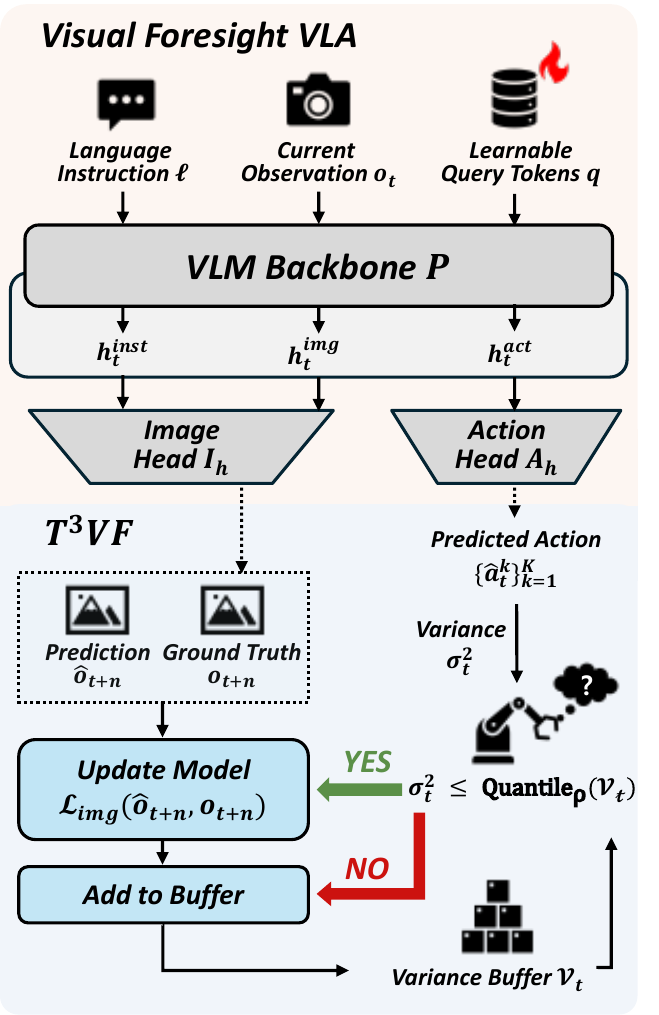}
    \caption{Overall framework of \proposed. Built on standard VF-VLA (upper), \proposed~(lower) generates $K$ action samples at each step and computes their variance $\sigma_t^2$. When $\sigma_t^2$ falls below the $\rho$ percentile of the variance buffer $\mathcal{V}_t$, the predicted-attained pair $(\hat{o}_{t+n}, o_{t+n})$ is used for test-time training. Regardless, $\sigma_t^2$ is added to $\mathcal{V}_t$.}
    \label{fig:framework}
\end{figure}

Building on this intuition, we introduce our method \proposed~(Test-Time Training for Visual Foresight VLA). The overall framework and detail algorithms are presented in Fig. \ref{fig:framework} and Algorithm \ref{alg:proposed}, respectively. At test time, we accumulate predicted-attained pairs along the trajectory of an episode into a set $\mathcal{B}$. Once $|\mathcal{B}|$ reaches a predefined batch size $B$, we apply a single update step that minimizes
\begin{equation}\label{eqa:4}
\mathcal{L}_{\mathrm{TTT}} \;=\; \frac{1}{B}\sum_{(\hat{o}_{t+n},\, o_{t+n})\in \mathcal{B}} \mathcal{L}_{\mathrm{img}}(\hat{o}_{t+n},\, o_{t+n}),
\end{equation}
where $\mathcal{L}_{\mathrm{img}}$ is the same image loss used during training. The update is applied only to $q$,\footnote{Which subset of parameters to update is depends on each VF-VLA implementation. We choose $q$ in our setting because it is the smallest module that participates in the image prediction pathway.} while the rest of the model parameters are kept frozen throughout test time. Since this update proceeds in parallel with execution and does not add auxiliary modules, the additional overhead is reduced.


\vspace{-0.5ex}
\subsection{Adaptive Update Filtering}\label{subsec:filter}
\vspace{-0.5ex}
However, applying test-time training of Sec.~\ref{subsec:ttt} requires addressing the following two practical challenges. First, \textbf{the source of a large prediction error cannot be identified from its magnitude alone.} A predicted-attained pair can exhibit a large discrepancy from a genuinely inaccurate visual prediction or from action-side error despite a correct prediction. Therefore indiscriminate test-time training risks canceling the gains from the former with the harm from the latter. Second, \textbf{a fixed threshold over proxy does not generalize across or within episodes.} Even with a reliable proxy for distinguishing the two cases, a fixed threshold is inappropriate to the test-time setting, because the difficulty varies both within and across episodes. Within an episode, easier segments allow too many update and may inject noise into the parameters, while harder segments allow almost none and suppress learning. The same effect manifests across episodes respectively, where overall easy episodes oversaturate updates and overall hard episodes receive almost no supervision.

 
We address the first challenge by using the variance of the action at each step as a proxy metric. Concretely, at step $t$ we draw $K$ samples $\{\hat{a}_t^{(k)}\}_{k=1}^{K}$ and define
\begin{equation}
\bar{a}_t \;=\; \frac{1}{K}\sum_{k=1}^{K} \hat{a}_t^{(k)}, \qquad \sigma_t^2 \;=\; \frac{1}{K}\sum_{k=1}^{K} \big\|\hat{a}_t^{(k)} - \bar{a}_t\big\|_2^2,
\end{equation}
where $\sigma_t^2$ is the squared L2 deviation averaged over the $K$ action samples. A low $\sigma_t^2$ indicates that the model is internally consistent with the action it intends to execute, so that any large prediction error at step $t$ is more plausibly attributable to the visual pathway. Therefore, it is a useful signal for updating $q$. In addition, this skip mechanism can be conducted at the moment $\hat{a}_t$ is produced, regardless of whether the paired $o_{t+n}$ is yet available, and the $K$ samples can be obtained from a single forward of the backbone followed by parallel decoding through the action head. Therefore, the additional cost is relatively less than calculating predicted-attained pairs through $\mathcal{L}_{\mathrm{img}}$.
 
We address the second challenge by replacing a fixed threshold with a running window of recent variance values. Let $\mathcal{V}_t = \{\sigma_{t'}^2 : t' \in \mathcal{W}_t\}$ be the buffer of variances collected over the most recent $|\mathcal{V}|$ steps, where $\mathcal{W}_t$ denotes the corresponding step indices. We include step $t$ in predicted-attained set $\mathcal{B}$ of Eq.~\ref{eqa:4} if and only if
\begin{equation}
\sigma_t^2 \;\le\; \mathrm{Quantile}_{\rho}(\mathcal{V}_t),
\end{equation}
where $\rho \in (0,1)$ is the percentile threshold. Because this criterion is based on relative ranking within the recent window rather than on an absolute cutoff, it naturally adapts to the scale of variance in each episode and keeps the overall frequency of accepted steps stable across episodes of differing difficulty. Combined with the test-time training procedure of Sec.~\ref{subsec:ttt}, this filtering completes T$^3$VF by turning indiscriminately occurring predicted-attained pairs into reliable supervision for the visual prediction pathway.

\vspace{-1.5ex}
\section{Experiments} \label{sec:experiment}
\vspace{-0.5ex}
\subsection{Setup}
\vspace{-0.5ex}
We evaluate \proposed~using Mantis \citep{yang2025mantis} as the baseline, a representative VF-VLA. We evaluate on LIBERO-Plus \citep{fei2025libero} following its standard evaluation protocol on the seven perturbation dimensions and report average success rates.
 \begin{table*}[!t]
\centering
\small
\setlength{\tabcolsep}{4pt}
\renewcommand{\arraystretch}{0.95}
\caption{Main results on LIBERO-Plus. Success Rate (\%) under two settings.}
\label{tab:main}
\begin{tabular}{ll *{7}{c} c}
\toprule
\textbf{Setting} & \textbf{Model} 
& \textbf{Robot} & \textbf{Language} & \textbf{Noise} & \textbf{Layout} 
& \textbf{Background} & \textbf{Camera} & \textbf{Light}
& \textbf{Avg} \\
\midrule
\multirow{3}{*}{w/ Perturbed Train}
  & Mantis             & 29.0 & 47.8 & 47.4 & 42.3 & 60.3 & 50.5 & 67.8 & 49.3   \\
  & Mantis + \proposed & 31.8 & 49.2 & 48.2 & 44.9 & 63.0 & 55.3   & 72.4   & 52.1   \\
  & $\boldsymbol{\Delta}$          & \textbf{+1.8} & \textbf{+1.4} & \textbf{+0.8} & \textbf{+2.6} & \textbf{+2.7} & \textbf{+4.8}   & \textbf{+4.6}   & \textbf{+2.8}   \\
\midrule
\multirow{3}{*}{w/o Perturbed Train}
  & Mantis             & 15.7 & 41.8 & 45.9 & 45.1 & 28.9 & 39.2 & 62.5 & 39.8 \\
  & Mantis + \proposed & 16.5 & 42.6 & 44.8 & 45.4 & 28.7 & 41.5   & 62.3   & 40.3   \\
  & $\boldsymbol{\Delta}$          & \textbf{+0.8} & \textbf{+0.8} & \textbf{-1.1} & \textbf{+0.3} & \textbf{-0.2} & \textbf{+2.3}   & \textbf{-0.2}   & \textbf{+0.5}   \\
\bottomrule
\end{tabular}
\end{table*}
We consider two settings: "w/ Perturbed Train" uses a model fine-tuned on LIBERO-Plus, partially adapted to the perturbations; "w/o Perturbed Train" uses the official LIBERO checkpoint, which is fully OOD at evaluation. Within each setting, we compare the base model with and without \proposed. For complete experimental setup, please refer to Appendix~\ref{app:setup}.

\vspace{-1ex}
\subsection{Main Results}
\vspace{-0.5ex}



Table~\ref{tab:main} shows that applying \proposed~consistently improves the overall average success rate in both settings. This pattern is in line with the motivation discussed in Sec.~\ref{sec:introduction}. \proposed~targets the visual prediction pathway with supervision extracted from predicted-attained image pairs, mitigating the dual-stage vulnerability of VF-VLA, and the resulting improvement in visual prediction propagates to action generation through their dependence. The relatively smaller improvement in the w/o Perturbed Train setting may reflect that the base model has not been adapted to the perturbation factors and therefore absorbs less supervision signal. However, \proposed~still yields a positive improvement.
\vspace{-1ex}
\subsection{Ablation Study}
\vspace{-0.5ex}
To isolate the contribution of each component of \proposed, we conduct an ablation study on Robot perturbation\footnote{Robot perturbation alters the initial state, affecting both the visual and action, and is therefore the most challenging OOD.} in the w/ Perturbed Train setting. Starting from the base model, we incrementally add (i) the test-time training of Sec.~\ref{subsec:ttt} without any filtering, (ii) the action variance filter of Sec.~\ref{subsec:filter} with a fixed threshold, and (iii) the adaptive variance buffer of Sec.~\ref{subsec:filter}, which together form \proposed. Table~\ref{tab:ablation} reports the success rate at each step.
\begin{table}[H]
\centering
\small
\setlength{\tabcolsep}{4pt}
\renewcommand{\arraystretch}{0.95}
\caption{Ablation of \proposed~components.}
\label{tab:ablation}
\begin{tabular}{cccc c}
\toprule
\textbf{TTT} & \textbf{Var.~Filter} & \textbf{Adap.~Buffer} & & \textbf{Success Rate} \\
\midrule
\textcolor{xmark_red}{\xmark} & \textcolor{xmark_red}{\xmark} & \textcolor{xmark_red}{\xmark} & & 29.0  \\
\textcolor{checkmark_green}{\cmark} & \textcolor{xmark_red}{\xmark} & \textcolor{xmark_red}{\xmark} & & 29.8  \\
 \textcolor{checkmark_green}{\cmark} & \textcolor{checkmark_green}{\cmark} &  \textcolor{xmark_red}{\xmark} & & 28.6  \\
\textcolor{checkmark_green}{\cmark} & \textcolor{checkmark_green}{\cmark} & \textcolor{checkmark_green}{\cmark} & & \textbf{31.8}  \\
\bottomrule
\end{tabular}
\end{table}
Adding the test-time training procedure on the base model improves the success rate, confirming that predicted-attained image pairs carry useful supervision even when applied indiscriminately. However, the fixed-threshold filter slightly degrades the success rate, indicating that an absolute cutoff cannot consistently separate informative from noisy steps, as discussed in Sec.~\ref{subsec:filter}. Replacing the fixed threshold with the adaptive variance buffer yields the largest improvement. This supports that our relative ranking criterion makes the proxy reliable.
\vspace{-1ex}
\subsection{Efficiency Analysis}
\begin{figure}[H]
\centering
    \includegraphics[width=\linewidth]{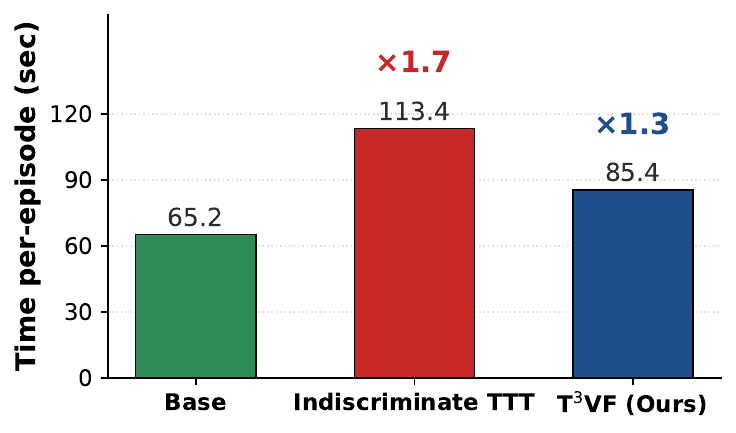}
    \caption{Efficiency Comparison across the settings.}
    \label{fig:efficiency}
\end{figure}

\vspace{-0.5ex}
Fig.~\ref{fig:efficiency} reports the average time per-episode on Robot perturbation setup, comparing the base model with two test-time training variants. The first variant is an indiscriminate test-time training that updates at every step without filtering, and the second is the full \proposed.
Indiscriminate test-time training increases the per-episode time to approximately 1.7 times the base time, reflecting the cost of computing $\mathcal{L}_{\mathrm{img}}$ and parameter update at every step. \proposed~reduces this to approximately 1.3 times through adaptive update filter, which triggers updates only on a fraction of the steps. Combined with the results in Fig. \ref{fig:efficiency} and the ablation in Table~\ref{tab:ablation}, \proposed~achieves performance improvement of test-time training while reducing the overhead relative to the indiscriminate variant.
\vspace{-1.5ex}
\section{Discussion} \label{sec:discussion}
\vspace{-0.5ex}
We position \proposed~as a lightweight approach to mitigate the OOD vulnerability of VF-VLA at test time, rather than a dominant solution. Although gains seem incremental relative to the additional inference cost, \proposed~does not require additional training pipelines and operates without auxiliary modules, external reward signals, and offline adaptation. In addition, Its overhead is further contained by using fast skip criterion and updating only the smallest module that participates in the visual prediction pathway. Within these constraints, \proposed~offers a practical option for adapting VF-VLA at test-time without architectural modification. 
\vspace{-1.5ex}
\section{Conclusion} \label{sec:conclusion}
We presented~\proposed, a test-time training approach that addresses the amplified OOD vulnerability of VF-VLA by leveraging predicted-attained image pairs as a self-supervision signal, complemented by an adaptive update filter. \proposed~offers a practical option for partially mitigating OOD shifts in VF-VLA without architectural modification.

\section*{Impact Statement}

This paper presents work whose goal is to advance the field of Machine
Learning. There are many potential societal consequences of our work, none
which we feel must be specifically highlighted here.


\bibliography{tttvfvla}
\bibliographystyle{icml2026}

\newpage
\appendix
\onecolumn
\section{Complete Related Work} \label{app:related_work}

\subsection{Visual Foresight VLA}

Visual Foresight VLA refer to VLA models that predict a future image to be reached by the robot from the current observation and language instruction, and condition action generation on this prediction. These models can be organized into two groups depending on whether they predict the image itself or its latent features. The first group formulates the prediction of future visual information and actions within a unified autoregressive token space, in which image, text, and action are treated as sequences on a shared discrete vocabulary. In this formulation, future images and actions are generated autoregressively within a single sequence \citep{zhao2025cot, zhang2025up, wang2025unified, cen2025worldvla}. The second group emerges as a refinement of the first, predicting future visual information as compressed latent features rather than discrete image tokens, or implicitly aligning latent representations with future visual states. This approach avoids the redundancy and computational overhead inherent to autoregressive pixel-token generation while still providing dense visual supervision \citep{yang2025mantis, zhangdreamvla}. Despite these differences, both groups share a common structure in which action generation is conditioned on predicted visual information, and consequently both the visual prediction stage and the action generation stage are exposed to OOD shifts leading to amplified vulnerability under such shifts.

\subsection{OOD Evaluation in VLA}
Several recent studies evaluate the out-of-distribution robustness of VLA models by injecting controlled perturbations into existing manipulation benchmarks \citep{pumacay2024colosseum, fei2025libero}. These studies consistently report that existing VLA models, despite achieving high success rates under in-distribution evaluation, are uniformly vulnerable once such perturbations are introduced.

\subsection{Test-Time Training in VLA}

Several recent studies attempt to adapt VLA models during deployment through test-time reinforcement learning \citep{bai2025evolve, liu2026fly}. They commonly rely on a learned progress estimator that replaces oracle reward signals and update the policy online via reinforcement learning. These approaches differ from ours in that they require training a separate reward model and incur substantial computational overhead from online reinforcement learning, and are designed as generic frameworks for VLA models.

\section{Complete Experimental Setup} \label{app:setup}

The hyperparameters of \proposed~are set to $n = 4$ for the prediction gap, $B = 4$ for the test-time training batch size, $K = 5$ for the action rollout count used in variance estimation, $|\mathcal{V}| = 10$ for the variance buffer size, and $\rho = 0.3$ for the percentile threshold. The base Mantis model serves as the comparison baseline within each setting. 

\begin{algorithm}[tb]
\caption{T$^3$VF: Test-Time Training Visual Foresight VLA}
\label{alg:proposed}
\begin{algorithmic}[1]
\STATE {\bfseries Input:} Trained VF-VLA $\{P, I_h, A_h, q\}$, instruction $l$, gap $n$, batch size $B$, sample count $K$, buffer size $|\mathcal{V}|$, percentile $\rho$, max step $T$
\STATE $\mathcal{B} \gets \emptyset$, $\mathcal{V} \gets \emptyset$
\FOR{$t = 1$ {\bfseries to} $T$}
   \STATE $(h_t^{\mathrm{inst}}, h_t^{\mathrm{img}}, h_t^{\mathrm{act}}) \gets P(l, o_t, q)$
   \STATE $\{\hat{a}_t^{(k)}\}_{k=1}^{K} \sim A_h(h_t^{\mathrm{act}})$
   \STATE $\bar{a}_t \gets \tfrac{1}{K}\sum_k \hat{a}_t^{(k)}$, \; $\sigma_t^2 \gets \tfrac{1}{K}\sum_k \|\hat{a}_t^{(k)} - \bar{a}_t\|_2^2$
   \IF{$|\mathcal{V}| < |\mathcal{V}|_{\max}$}
      \STATE $\mathcal{V}.\mathrm{append}(\sigma_t^2)$
      \STATE Execute $\bar{a}_t$
      \STATE \textbf{continue}
   \ELSE
      \STATE $\mathcal{V}.\mathrm{pop}()$, $\mathcal{V}.\mathrm{append}(\sigma_t^2)$
   \ENDIF
   \STATE Execute $\bar{a}_t$
   \IF{$\sigma_t^2 \le \mathrm{Quantile}_{\rho}(\mathcal{V})$}
      \STATE $\hat{o}_{t+n} \gets I_h([h_t^{\mathrm{inst}}, h_t^{\mathrm{img}}], o_t)$
      \STATE $\mathcal{B} \gets \mathcal{B} \cup \{(\hat{o}_{t+n}, o_{t+n})\}$ once $o_{t+n}$ available
   \ENDIF
   \IF{$|\mathcal{B}| = B$}
      \STATE $q \gets q - \alpha \nabla_q \mathcal{L}_{\mathrm{TTT}}$ \quad \{TTT update\}
      \STATE $\mathcal{B} \gets \emptyset$
   \ENDIF
   \STATE {\bfseries if} task done {\bfseries then break}
\ENDFOR
\end{algorithmic}
\end{algorithm}

\end{document}